\documentclass[final]{cvpr}

\usepackage{times}
\usepackage{epsfig}
\usepackage{graphicx}
\usepackage{amsmath}
\usepackage{amssymb}
\usepackage{color}

\usepackage{cuted}
\usepackage{capt-of}
\usepackage{overpic}
\usepackage{verbatim}
\usepackage{multirow}

\usepackage{booktabs}

\usepackage{lipsum}

\usepackage{booktabs}
\DeclareMathAlphabet{\mathscrbf}{OMS}{mdugm}{b}{n}

\usepackage[pagebackref=true,breaklinks=true,colorlinks,bookmarks=false]{hyperref}

\def\cvprPaperID{1128} 
\def\confYear{CVPR 2021}

\begin{document}

\title{Recurrent Multi-view Alignment Network for Unsupervised Surface Registration}

\author{
	{\large Wanquan Feng\textsuperscript{1} \quad Juyong Zhang\textsuperscript{1}\thanks{Corresponding author} \quad Hongrui Cai\textsuperscript{1}  \quad Haofei Xu\textsuperscript{1} \quad Junhui Hou\textsuperscript{2} \quad Hujun Bao\textsuperscript{3}}
	\\
	{\normalsize \textsuperscript{1}University of Science and Technology of China \quad \textsuperscript{2}City University of Hong Kong \quad \textsuperscript{3}Zhejiang University}
	\\ {\tt\footnotesize \{lcfwq@mail., juyong@, hrcai@mail., xhf@mail.\}ustc.edu.cn \hspace{1 mm} jh.hou@cityu.edu.hk \hspace{1 mm} bao@cad.zju.edu.cn}
}

\twocolumn[{
\maketitle
\vspace*{-12mm}
\begin{center}
\begin{overpic}
	[width=\textwidth]{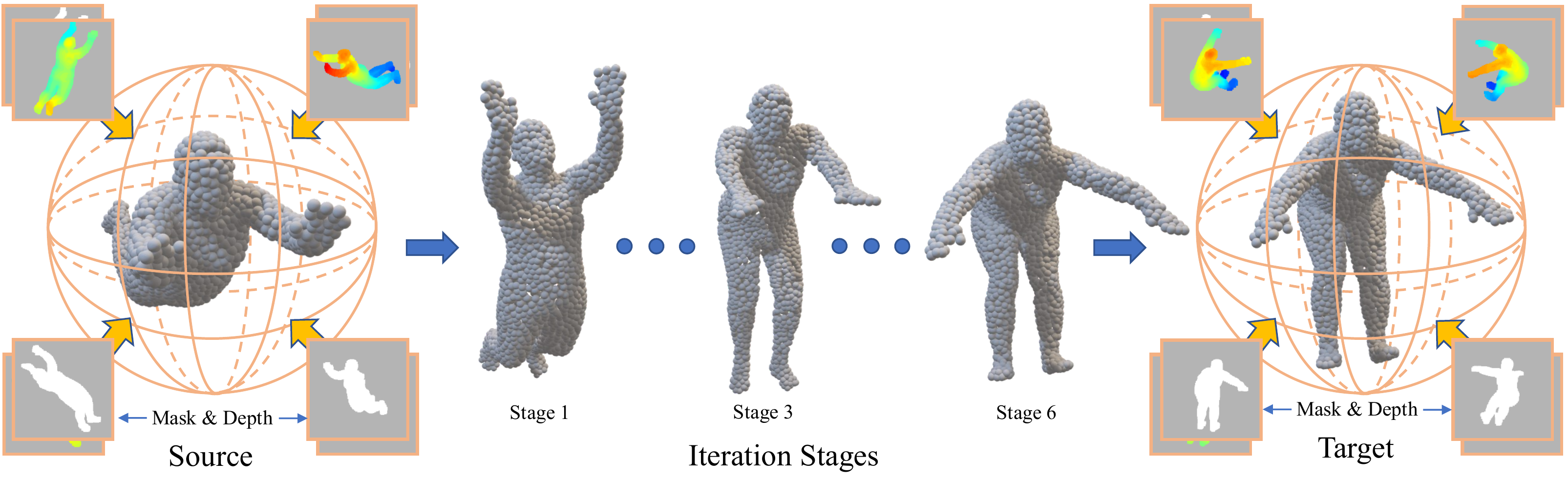}	
\end{overpic}
\end{center}
\vspace*{-5mm}
\captionof{figure}{We propose RMA-Net for non-rigid registration. With a recurrent unit, the network iteratively deforms the input surface shape stage by stage until converging to the target. RMA-Net is totally trained in an unsupervised manner by aligning the source and target shapes via our proposed multi-view 2D projection loss.}
\label{fig:teaser}
\vspace*{2mm}
}]

{

\renewcommand{\thefootnote}%

{\fnsymbol{footnote}}

\footnotetext[1]{*Corresponding author}

}


\begin{abstract}

\vspace{-3mm}
Learning non-rigid registration in an end-to-end manner is challenging due to the inherent high degrees of freedom and the lack of labeled training data. In this paper, we resolve these two challenges simultaneously. First, we propose to represent the non-rigid transformation with a point-wise combination of several rigid transformations. This representation not only makes the solution space well-constrained but also enables our method to be solved iteratively with a recurrent framework, which greatly reduces the difficulty of learning. Second, we introduce a differentiable loss function that measures the 3D shape similarity on the projected multi-view 2D depth images so that our full framework can be trained end-to-end without ground truth supervision. Extensive experiments on several different datasets demonstrate that our proposed method outperforms the previous state-of-the-art by a large margin. The source codes are available at \href{https://github.com/WanquanF/RMA-Net}{https://github.com/WanquanF/RMA-Net}.
\end{abstract}


\vspace{-5mm}

\section{Introduction}
\label{Introduction}


Surface registration, which aims to find the spatial transformation and correspondences between two surfaces, is a fundamental problem in computer vision and graphics. It has wide applications in many fields such as 3D reconstruction~\cite{NewcombeIHMKDKSHF11,NewcombeFS15}, tracking~\cite{DBLP:conf/cvpr/Qian0WT014, DBLP:journals/pami/VedulaBRCK05} and medical imaging~\cite{DBLP:conf/iccv/ZhaoDCX19, DBLP:conf/miccai/SentkerMW18}. According to the type of transformation that deforms the source surface to the target, registration algorithms can be categorized as rigid~\cite{DBLP:journals/pami/BeslM92, DBLP:conf/cvpr/AokiGSL19,DBLP:conf/iccv/WangS19,DBLP:journals/corr/abs-1905-04153,DBLP:conf/cvpr/YewL20,DBLP:conf/cvpr/ChoyDK20,DBLP:conf/cvpr/HuangM020} and non-rigid~\cite{DBLP:journals/pami/MyronenkoS10, DBLP:journals/corr/abs-1906-03039, DBLP:journals/corr/abs-1904-01428, 0A, DBLP:conf/cvpr/LiuQG19} methods. Rigid registration estimates a global rotation and translation that aligns two surfaces, while non-rigid registration has much more freedoms, and thus is more complex and challenging.

Traditional optimization based methods~\cite{DBLP:journals/pami/BeslM92, DBLP:journals/pami/MyronenkoS10} usually handle this problem by iteratively alternating a correspondence step and an alignment step. The correspondence step is to find corresponding points between the source and target surfaces, while the alignment step estimates the transformation based on the current correspondences. However, it is not easy to construct reliable correspondences solely based on heuristics (\eg, nearest neighbor search) or handcrafted features like SHOT~\cite{TombariSS10} and FPFH~\cite{DBLP:conf/icra/RusuBB09}.

Recently, learning based methods have demonstrated promising results by leveraging the strong representation learning abilities of neural networks, but most of them are limited to rigid registration~\cite{DBLP:conf/cvpr/AokiGSL19,DBLP:conf/iccv/WangS19,DBLP:journals/corr/abs-1905-04153,DBLP:conf/cvpr/YewL20,DBLP:conf/cvpr/ChoyDK20,DBLP:conf/cvpr/HuangM020}. Only a few learning based non-rigid registration methods~\cite{DBLP:journals/corr/abs-1904-01428, DBLP:journals/corr/abs-1906-03039} exists and they are only applicable to small-scale non-rigid deformations that are synthesized by the thin plate spline (TPS)~\cite{DBLP:journals/pami/Bookstein89} transformation. It is not straightforward to design a learning based non-rigid registration framework that can produce very accurate results.

The challenges of learning non-rigid registration lie in two aspects. First, unlike a single global rigid transformation, the much greater freedoms in non-rigid registration increase the difficulty of network training. For example, traditional methods~\cite{DBLP:conf/cvpr/AmbergRV07, DBLP:journals/pami/MyronenkoS10} estimate a local transformation for every points on the surface. To alleviate this issue, one possible direction is to adopt the deformation graph~\cite{DBLP:journals/tog/SumnerSP07} representation, which reduces the complexity from each surface point to each graph node. However, it is not convenient to pass the shape-dependent deformation graphs to neural network due to their different number of graph nodes and topologies. Second, the lack of labeled data restricts the training of non-rigid registration networks. It is non-trivial to obtain dense non-rigid correspondences from real large-scale data to serve as direct supervision. An alternative is to train the network in an unsupervised fashion. However, existing metrics (\eg, chamfer distance and earth-mover distance) for shape similarity measurement are not effective enough to drive the network to learn the correct solution.



To tackle these problems, we first propose a new non-rigid representation that is suitable for network learning. Specifically, we represent the non-rigid transformation as a point-wise combination of $K$ rigid transformations, where $K$ is much smaller than the number of surface points. This is because the non-rigid deformation of the surface can be well approximated by the combination of several rigid transformations. Such a representation not only enables our method to be able to approximate arbitrary non-rigid transformation but also makes the solution space well-constrained. To learn such a representation, we design a recurrent neural network architecture to estimate the combination weights and each rigid transformation iteratively. At each iteration, the network only needs to estimate a single rigid transformation and the skinning weight for each point which represents how important the rigid transformation influences this point. 
Although the iterative strategy has been adopted in neural networks for registration~\cite{DBLP:conf/cvpr/AokiGSL19,DBLP:journals/corr/abs-1908-07906}, they are mainly used for rigid registration, while our proposed iterative method is well-designed for non-rigid registration.

We further propose a multi-view loss function to train the model in a self-supervised manner. Concretely, we project the 3D surface to multi-view 2D depth images and measure the visual similarity of source and target by their depths and masks. The intuitive idea is that their projected depth maps and masks should be identical if two surfaces are fully registered. Moreover, we adopt a soft rasterization method to render the point clouds to depths map and masks, which makes our loss term to be differentiable. The proposed shape similarity loss is adopted in our self-supervised learning method, and it outperforms the commonly used Chamfer Distance and Earth Mover's distance.


We conduct extensive experiments on different object types (body, face, cats, dogs, and ModelNet40 dataset). Our proposed RMA-Net not only outperforms previous state-of-the-art methods by a large margin but also works well for large-scale non-rigid registration tasks. Extensive ablation studies also validate the effectiveness of each component in our proposed method.


\section{Related Works}
\label{Related}

{\noindent \bf Rigid Registration.} A popular rigid registration method is the Iterative Closest Point (ICP)~\cite{DBLP:journals/pami/BeslM92} algorithm, which searches the correspondences and estimates the transformation alternatively to solve the problem. Some ICP variants~\cite{DBLP:journals/cgf/BouazizTP13, DBLP:conf/3dim/RusinkiewiczL01, DBLP:conf/rss/SegalHT09, DBLP:conf/icra/ServosW14} have been proposed to improve its robustness noises, outliers and incomplete scans. 
On the other hand, some global optimization based methods~\cite{DBLP:journals/cacm/FischlerB81,DBLP:conf/icra/RusuBB09,DBLP:journals/ijrr/RosenCBL19,  DBLP:conf/isrr/IzattDT17, DBLP:journals/pami/YangLCJ16} have been proposed to search for a global optimum while at the cost of slow computation speed.

Recently, learning-based methods have also shown promising results. 3DMatch~\cite{DBLP:conf/cvpr/ZengSNFXF17} and 3DFeatNet~\cite{DBLP:conf/eccv/YewL18} learn local patch descriptors instead of hand-crafted features to construct correspondences. PointNetLK~\cite{DBLP:conf/cvpr/AokiGSL19} and PCRNet~\cite{DBLP:journals/corr/abs-1908-07906} extract global features of the input point clouds and iteratively regresses the rigid transformation. Deep Closest Point (DCP)~\cite{DBLP:conf/iccv/WangS19} improves the feature extraction and correspondence prediction stages, and obtains the rigid transformation through SVD decomposition. PR-Net~\cite{DBLP:conf/nips/WangS19} and RPM-Net~\cite{DBLP:conf/cvpr/YewL20} improve the robustness to outliers and partial visibility.
Recently, ~\cite{DBLP:conf/cvpr/HuangM020} proposes a semi-supervised approach based on feature-metric projection error, which also demonstrates robustness to noise, outliers, and density difference. Some of these methods adopt recurrent structures to update the transformation iteratively while only work for rigid transformation. Different from our approach, most of these methods are trained in a supervised manner or with Chamfer distance loss.


{\noindent \bf Non-Rigid Representation.} One widely used non-rigid representations is to estimate a local transformation for every point on the source surface~\cite{DBLP:conf/cvpr/AmbergRV07, DBLP:journals/pami/MyronenkoS10}. Although such representation can model complex non-rigid deformations, the large number of variables increases the difficulty of network training. Another type is based on deformation graph~\cite{DBLP:journals/tog/SumnerSP07}. Given an input surface, the deformation graph is constructed by sampling graph nodes on the surface. By defining transformation variables on graph nodes, the surface can be deformed based on the skinning weights which tie between points of the surface and nodes. However, different surfaces have deformation graphs with a different number of graph nodes, topologies and skinning weights. Such a shape-dependent representation makes it difficult to be passed into the neural network. Unlike existing methods, our proposed non-rigid representation is flexible and applicable for different shape types, which can also be easily learned with our proposed recurrent framework.

{\noindent \bf Non-Rigid Registration.} Several previous non-rigid registration methods are based on thin plate spline functions~\cite{DBLP:journals/cviu/ChuiR03, DBLP:journals/pami/JianV11, DBLP:journals/sigpro/ChenMYMZ15, DBLP:journals/pr/YangOF15} and the local affine transformations~\cite{DBLP:conf/cvpr/AmbergRV07,  DBLP:journals/cgf/LiSP08, DBLP:conf/cvpr/YaoDXZ20}. N-ICP~\cite{DBLP:conf/cvpr/AmbergRV07} defines a rigid transformation for each point to model the non-rigid deformation. Embedded deformation-based methods~\cite{DBLP:journals/cgf/LiSP08, DBLP:journals/tog/SumnerSP07, DBLP:conf/cvpr/YaoDXZ20} model the non-rigid motion as transformations defined on a deformation graph. Coherent point drift (CPD)~\cite{DBLP:journals/pami/MyronenkoS10} is based on the Gaussian mixture model (GMM), which implicitly encodes the unknown correspondences between points and minimizes the negative log-likelihood function using the Expectation-Maximization (EM) algorithm. Some other GMM-based methods~\cite{DBLP:journals/pami/KolesovLSVT16, DBLP:journals/tip/Ma0Y16, DBLP:conf/wacv/GolyanikTRS16} are also proposed later. Very recently, BCPD~\cite{0A} formulates CPD in a Bayesian framework, which guarantees the convergence and reduces the computation time.

At present, learning based surface registration methods are mainly designed for rigid registration, while only a few exist for non-rigid registration. CPD-Net~\cite{DBLP:journals/corr/abs-1906-03039} extracts features from input point cloud pairs with PointNet backbone and predicts the point-wise displacement directly, which is trained in an unsupervised manner with the Chamfer loss. PR-Net~\cite{DBLP:journals/corr/abs-1904-01428} adopts a voxel-based strategy to extract shape correlation tensor and predict the control points of thin plate spine. Training is supervised with a GMM loss. FlowNet3D~\cite{DBLP:conf/cvpr/LiuQG19} estimates the scene flow from a pair of consecutive point clouds, which is trained with ground truth supervision. Different from previous methods, we propose a recurrent framework to learn non-rigid registration in an unsupervised manner, and also a new loss function is proposed to drive the learning process.


{\noindent \bf Metrics for Shape Similarity.} The key to a successful unsupervised learning framework is to design suitable loss functions. The Chamfer Distance (CD) is a general evaluation metric in related area, including surface generation~\cite{DBLP:conf/cvpr/FanSG17,DBLP:journals/corr/abs-1802-05384}, point set registration~\cite{DBLP:journals/corr/abs-1906-03039, DBLP:conf/cvpr/HuangM020} and so on. As CD relies on the closet point distance, it is sensitive to the detailed geometry of outliers~\cite{DBLP:conf/cvpr/TatarchenkoRRLK19}. Earth Mover's distance (EMD) is another metric that is commonly used to compute the distance between two surfaces~\cite{DBLP:conf/cvpr/FanSG17,DBLP:conf/cvpr/WangCMN19}, which can be formulated and solved as a transportation problem. As computing EMD is quite slow, some fast numerical methods~\cite{DBLP:conf/cvpr/FanSG17} have also been proposed.

Different from directly measuring the registration results in 3D, we propose to project the 3D shapes to a 2D plane and then measure their similarity. If two surface shapes match quite well, the rendered 2D images from different views will also be visually similar. For the 3D shape retrieval task, ~\cite{DBLP:journals/cgf/ChenTSO03} proposes a light field distance (LFD) to measure the visual similarity between two 3D models. Later, ~\cite{DBLP:journals/tog/GaoYQLRXX18} utilizes a neural network to learn this similarity metric for the guidance of deformation transfer. They all treat LFD as a non-differentiable module and thus can not be used as a loss term to supervise the model training. Compared with existing LFD approaches,	we design a differentiable rendering strategy to render the depths and masks from the given surface and camera parameters, which is similar to the method used in~\cite{DBLP:journals/corr/abs-1901-05567,TanZCZPT20}. In this way, our proposed loss is differentiable and can thus deform the surface accordingly.

\section{Proposed Model}
\label{Algorithm}

\subsection{Problem Definition}
\label{Overview}

Given a source point cloud $\mathcal{S} \in \mathbb{R}^{M \times 3}$ and a target point cloud $\mathcal{T} \in \mathbb{R}^{N \times 3}$, we aim to find a non-rigid transformation $\mathbf{\phi}: \mathbb{R}^{M \times 3} \rightarrow \mathbb{R}^{M \times 3}$, such that the deformed point cloud
\begin{equation}
\tilde{\mathcal{S}} = \mathbf{\phi}(\mathcal{S})
\label{eq:s'}
\end{equation}
is as close as possible to the target point cloud $\mathcal{T}$. The goal of this paper is to design a learning based framework to directly predict the non-rigid transformation $\phi$ with the source and target surfaces $\mathcal{S}$ and $\mathcal{T}$ as input.


\begin{figure*}[htb]
	\centering
	\includegraphics[width=2\columnwidth]{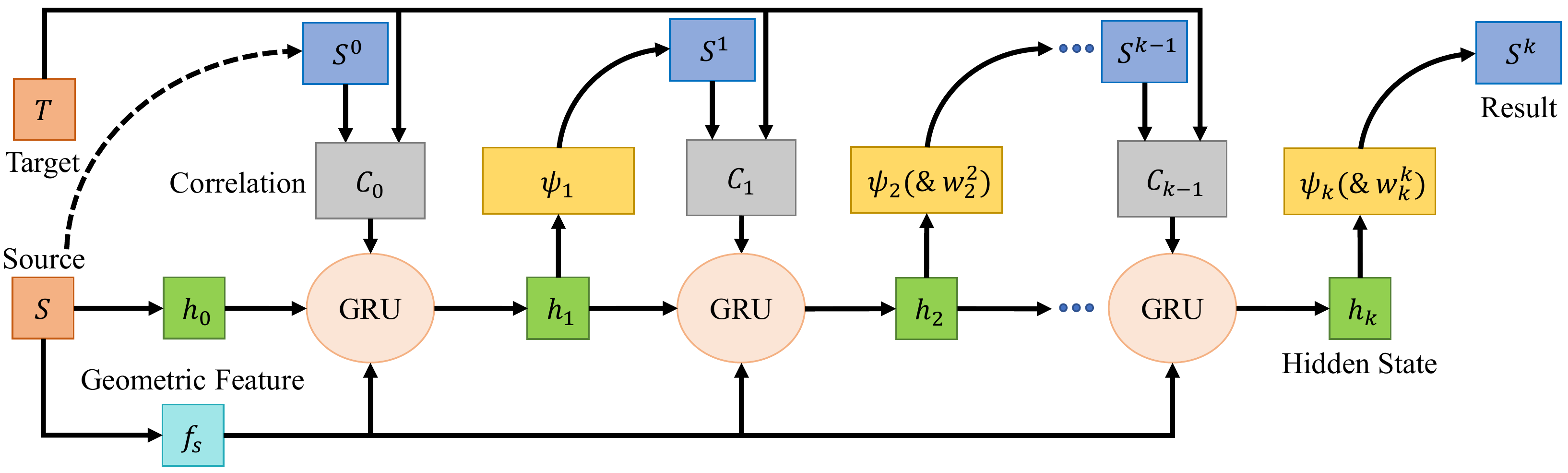}
	\caption{Illustration of RMA-Net. In the GRU-based framework, the hidden state is initialized (denoted as $h_{0}$) by extracting a feature from the source and updated (denoted as $h_{k}$ in the $k$-th stage) during the recurrent stages. In the $k$-th stage, we extract the correlation $\mathcal{C}_{k-1}$ of the current deformed surface $\mathcal{S}^{k-1}$ and target surface $\mathcal{T}$. The geometric feature $f_{s}$ of the source surface is extracted once, and it is concatenated together with the correlation as the input of updating unit in each stage. One rigid transformation $\mathbf{\psi}_{k}$ and its corresponding point-wise skinning weight $\mathbf{w}_{k}^{k}$ are regressed from $h_{k}$.}
	\label{fig:network}
    \vspace*{-5mm}
\end{figure*}

\subsection{Proposed Non-Rigid Representation}
\label{sec:representation}
We propose to represent the non-rigid transformation $\mathbf{\phi}$ with a point-wise combination of a series of rigid transformations $\{\mathbf{\psi}_{r}\}_{r=1}^{K}$:
\begin{equation}
\mathbf{\phi}(\mathcal{S}) = \sum_{r=1}^{K} \mathbf{w}_{r} \cdot \mathbf{\psi}_{r}(\mathcal{S}),
\label{eq:companation}
\end{equation}
where $\mathbf{w}_{r} \in \mathbb{R}^{M \times 1}$ is the point-wise skinning weight, the $\cdot$ denotes the point-wise multiplication, and $K$ is the number of rigid transformations that is much smaller than the number of points. 

For each point, we constrain the weights assigned to all rigid transformations to satisfy the condition that their sum equals to $1$:
\begin{equation}
\sum_{r=1}^{K} \mathbf{w}_{r}(i) = 1, \quad \forall i = 1, 2, \cdots, M.
\label{eq:weightsum1}
\end{equation}
It degenerates to rigid transformation when $K=1$, and it can represent non-rigid transformation when $K\geq 2$ as each surface point is influenced by more than one rigid transformation with different skinning weights. Its representation ability is gradually enhanced when $K$ becomes larger.

Compared with the deformation graph based representation, our method enjoys the following benefits:
\begin{itemize}
	\item Different from deformation graph that defines local transformation on graph nodes, our proposed model does not need to construct deformation graph for each specific surface as the rigid transformation $\mathbf{\psi}_{r}$ in Eq.~\eqref{eq:companation} is defined globally for all points on the surface.
	\item Different from the fixed skinning weights for a given surface and deformation graph, the skinning weights in Eq.~\eqref{eq:companation} are learned, and they can be adaptively adjusted according to different source and target surfaces.
\end{itemize}

Such a representation can not only express complex non-rigid representations, it can also be easily extended to rigid registration by removing the weights and changing the addition to multiplication in Eq.~\eqref{eq:companation}: $\mathbf{\phi}(\mathcal{S}) = \mathbf{\psi}_{K} \circ \mathbf{\psi}_{K-1} \circ \cdots \circ \mathbf{\psi}_{1}(\mathcal{S})$.

\subsection{Recurrent Update Framework}
\label{recurrent_update_strategy}

However, the number of variables in Eq.~\eqref{eq:companation} ($\{\mathbf{\psi}_{r}\}_{r=1}^{K}$ and $\{\mathbf{w}_{r}\}_{r=1}^{K}$) is still quite large, including $M\times K$ skinning weights and $6\times K$ rigid transformations. Thus, it may be not easy to predict all variables directly at the same time. To handle this problem, we propose a recurrent updating strategy to regress the rigid transformations and skinning weights in a stage-wise manner. At the $k$-th stage, the deformed point cloud is expressed as:
\begin{equation}
\begin{split}
\mathcal{S}^{k} = \sum_{r=1}^{k} \mathbf{w}_{r}^{k} \cdot \mathbf{\psi}_{r}(\mathcal{S}),
\label{eq:sk_formula}
\end{split}
\end{equation}
where the point-wise weight of the $r$-th rigid transformation at stage $k$ is denoted as $\mathbf{w}_{r}^{k}$ and we always keep the constraint $\sum\limits_{r=1}^{k} \mathbf{w}_{r}^{k} = \mathbf{1}$ to be satisfied.

Next we introduce how we recurrently obtain the gradually deformed point cloud $\{\mathcal{S}^{k}\}_{k=1}^{K}$. At the $1$-st stage, our network predicts a single rigid transformation $\mathbf{\psi}_{1}$ and we can obtain a transformed point cloud $\mathcal{S}^{1} = \mathbf{w}_{1}^{1} \cdot \mathbf{\psi}_{1}(\mathcal{S})$, where $\mathbf{w}_{1}^{1} \equiv \mathbf{1}$. At the $k$-th stage when $k \geq 2$, the network regresses $\mathbf{\psi}_{k}$ and $\mathbf{w}_{k}^{k}$. To satisfy the constraint in Eq.~\eqref{eq:weightsum1}, we can scale the predicted weights at previous stages ($\{\mathbf{w}_{r}^{k}\}_{r=1}^{k-1}$) with a factor of $\mathbf{1}-\mathbf{w}_{k}^{k}$, and thus the following updating formula can be obtained:
\begin{equation}
\begin{split}
\mathbf{w}_{r}^{k} = (\mathbf{1}-\mathbf{w}_{k}^{k}) \cdot \mathbf{w}_{r}^{k-1}, 1 \le r \le k-1. 
\label{eq:weight_recurrence_formula}
\end{split}
\end{equation}
It can be verified that Eq.~\eqref{eq:weightsum1} is satisfied at all stages according to the updating formula. Accordingly, we can derive the recurrence formula for the deformed point cloud at each stage:
\begin{equation}
\begin{split}
\mathcal{S}^{k} = (\mathbf{1}-\mathbf{w}_{k}^{k}) \cdot \mathcal{S}^{k-1} + \mathbf{w}_{k}^{k} \cdot \mathbf{\psi}_{k}(\mathcal{S}).
\label{eq:recurrence_formula}
\end{split}
\end{equation}

Based on the above formulations, our network only needs to predict one rigid transformation and the skinning weights at each stage, which greatly reduces the difficulty of learning. In this way, we iteratively deform the point cloud such that the sequence of deformed point clouds converges to the target gradually.

\section{RMA-Net and Loss}
\label{Network}

In this section, we give details of our proposed Recurrent Multi-view Alignment Network (RMA-Net) and the loss functions.

\subsection{Network Architecture}
\label{network_structure}
We adopt a recurrent network to update the deformed point cloud iteratively. The network includes two key components. First, deep features of the input source and target point clouds are extracted, and their correlations are computed by the dot-product operation. Second, a recurrent update module is used to implement the iteration process. Fig.~\ref{fig:network} provides an overview of our full framework.

\noindent{\bf Feature Extraction and Correlation Computation.}
Similar to~\cite{DBLP:conf/iccv/WangS19}, we first extract deep features from the input point clouds $\mathcal{S}^{k-1}$ and $\mathcal{T}$ with DGCNN~\cite{DBLP:journals/tog/WangSLSBS19} and Transformer~\cite{DBLP:conf/nips/VaswaniSPUJGKP17}. The features are of size $M \times C$ and $N \times C$, where $C$ is the feature channels. Then a correlation tensor is computed with the dot-product operation between every feature vector in these two features. The correlation size is $M \times N$. A top-K operation is next performed on the last dimension of the correlation, which removes the dependence on the number of target points $N$. The resulting correlation has the same dimension as the source feature. The concatenation of source feature, global target feature (average-pooling and expansion of the target feature), and the correlation is used during the update, denoted as $\mathcal{C}_{k}$. 


\noindent{\bf GRU-based Update.} 
We adopt a GRU~\cite{DBLP:conf/ssst/ChoMBB14} to implement the recurrent update process in Sec.~\ref{recurrent_update_strategy}. Although the overall architecture is similar to RAFT~\cite{DBLP:conf/eccv/TeedD20}, a recent work for optical flow estimation, there are several important differences. First, we focus on irregular point clouds other than the regular pixels, and thus the same architecture is not applicable here. Second, our key contribution is a new non-rigid representation, and the recurrent framework is only used for ease of optimization. Besides, the updating formula in Eq.~\eqref{eq:recurrence_formula} is clearly different from RAFT.



Before the first iteration, we also extract an initial hidden state $h_{0}$ and a geometric feature $f_{s}$ from the source point cloud with an additional DGCNN. At the $k$-th stage, we concatenate the geometric feature $f_{s}$ and the correlation $\mathcal{C}_{k-1}$ together $x_{k} =  [\mathcal{C}_{k-1},f_{s}]$ as the input of the update unit. The fully connect layers in GRU are replaced with MLPs. From $x_{k}$ and $h_{k-1}$, following~\cite{DBLP:conf/eccv/TeedD20}, the GRU obtains the updated hidden state $h_{k-1}$. Then, $h_{k}$ is passed through two MLPs to predict $\mathbf{w}_{k}^{k}$ and $\mathbf{\psi}_{k}$. The new deformed point cloud $\mathcal{S}^{k}$ can be obtained according to Eq.~\eqref{eq:recurrence_formula}.

\subsection{Loss Function}
\label{Loss}
\begin{figure}[t]
	\centering
	\includegraphics[width=1\columnwidth]{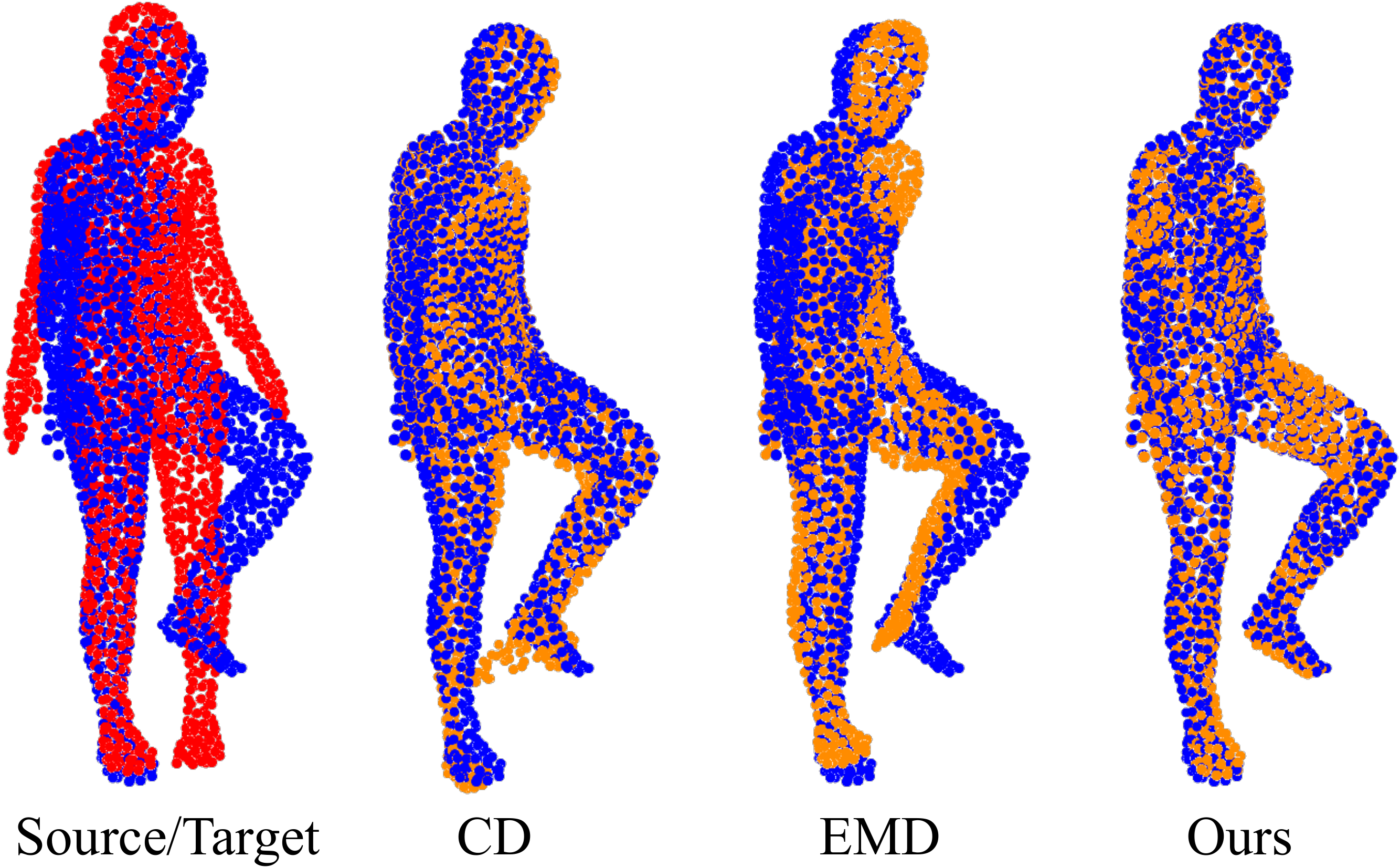}
	\caption{Comparison of different loss functions by over-fitting one sample pair. The results are shown in orange. From the source (red) to the target (blue), both CD and EMD loss terms tend to pull the left leg to the right leg, while our loss can successfully deform to the target shape.}
	\label{fig:loss_motivation}
    \vspace*{-5mm}
\end{figure}

The key to a successful unsupervised learning framework is to design suitable loss functions. Although Chamfer distance (CD) and Earth Mover's distance (EMD) are commonly used metrics to measure the distance of surface shapes, they depend on the closest point or the best transporting flow, causing the high failure rate when deforming the source surface to the target by minimizing CD or EMD. One example is shown in Fig.~\ref{fig:loss_motivation}, where we overfit the model on one pair with different loss functions. Both CD and EMD loss functions are unable to deform the source point cloud to the target accurately.

In this paper, instead of directly searching for the closest point or the best transporting flow, we construct the loss function based on the metric of shape similarity~\cite{DBLP:journals/cgf/ChenTSO03,DBLP:journals/tog/GaoYQLRXX18,DBLP:journals/corr/abs-1901-05567,TanZCZPT20}. Similar to the Light Field Descriptor (LFD)~\cite{DBLP:journals/cgf/ChenTSO03}, we project the 3D shapes onto multi-view 2D planes and measure the similarity between the deformed shape and the target via the projected 2D depth and mask images. To make the whole process trainable, we design a differentiable rendering method to render the point cloud to 2D depths and masks. Besides, we design regularization terms for the transformation variables and skinning weights. In the following, for convenience, we use $\tilde{\mathcal{S}}$ to denote the deformed shape $\mathcal{S}^{k}$ at stage $k$.

\begin{figure*}[t]
	\centering
	\includegraphics[width=2\columnwidth]{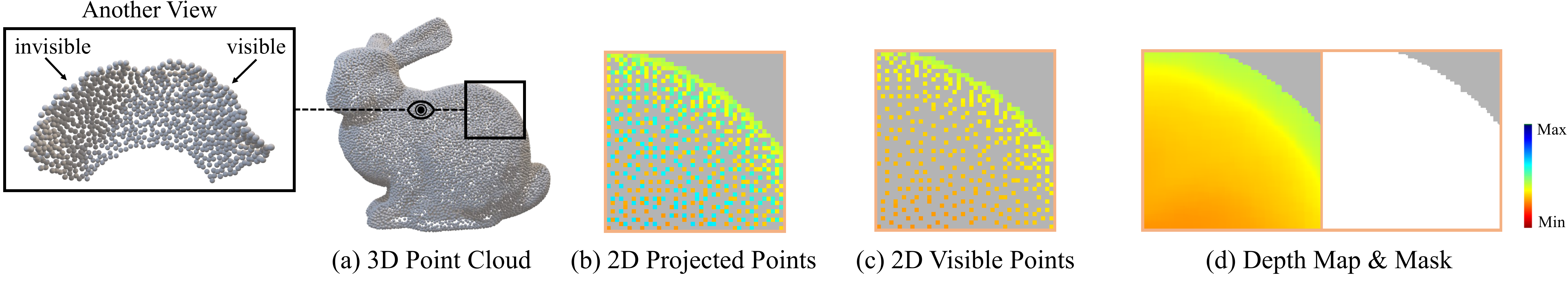}
	\caption{Illustration of the differentiable rendering process from the 3D point cloud to the 2D depths and masks. (a): the input point cloud. (b): Given the point cloud and camera, we project all the points to the front-view and set the value of the depth map as the z-value of the projected points. (c): We remove the invisible points around pixel $p_i$ based on the depth values of points projected in the $p_i$-centered window. (d): The depth value of $p_i$ is computed by a weighted average of the z-value of visible points projected in the window, and the mask of the object can also be recovered accordingly.}
	\label{fig:depthmap}
	\vspace{-3mm}
\end{figure*}

\noindent{\bf Depth Loss.}
For a point cloud $\mathcal{P}$ and a given viewing angle $v$, we transform $\mathcal{P}$ to the camera coordinate system as $\mathcal{P}_{v}$ and compute the depth map $\mathcal{D}(\mathcal{P}_{v})$. We first collect the points in $\mathcal{P}_v$ whose 2D projection are in the $k_{D} \times k_{D}$ window of pixel $p_i$ on $\mathcal{D}(\mathcal{P}_v)$, and define the set of these points as $\mathcal{N}(p_i)$. Then, the minimal and maximal z-value of $\mathcal{N}(p_i)$ are denoted as $\min_i, \max_i$. As shown in (b) and (c) of Fig.~\ref{fig:depthmap}, we remove the points whose z-value exceeds $(\min_i+\max_i)/2$ from $\mathcal{N}(p_i)$ since they may be from the invisible part, and we denote the visible part of $\mathcal{N}(p_i)$ as $\mathcal{V}(p_i)$. For $g_j \in \mathcal{V}(p_i)$, we set the weight $w_{ij}$ as:
\begin{equation}
\label{eq:depth_weight}
w_{ij}=\frac{\exp(-\rho_{ij}/\gamma)}{\sum\nolimits_{g_m\in \mathcal{V}(p_i)}\exp(-\rho_{im}/\gamma)},
\end{equation}
where $\rho_{ij}$ denotes the squared distance between $p_i$ and 2D projection of $g_j$, $\gamma$ controls the sharpness of the depth map. In this way, the depth value $d_i$ of pixel $p_i$ on $\mathcal{D}(\mathcal{P}_v)$ is computed by a weighted average of the z-value of points in $\mathcal{V}(p_i)$:
\begin{equation}
\label{eq::depth_cal}
d_i=\sum\nolimits_{g_j\in\mathcal{V}(p_i)}w_{ij}g_j^z,
\end{equation}
where $g_j^z$ denotes the z-value of points $g_j$.

For point cloud $\tilde{\mathcal{S}}$ and $\mathcal{T}$, we compute their depth as $\mathcal{D}(\tilde{\mathcal{S}}_v)$ and $\mathcal{D}(\mathcal{T})$. The loss between these paired depth maps is defined as
\begin{equation}
\label{eq:local_depth_loss}
\mathbf{L}_{\mathrm{depth}}(\tilde{\mathcal{S}},\mathcal{T})=\mathbb{E}_{v\sim V} \left\|\mathcal{D}(\tilde{\mathcal{S}}_v)-\mathcal{D}(\mathcal{T}_v)\right\|_2^2,
\end{equation}
where $V$ denotes the set of camera views. During back-propagation, the gradient $\nabla d_i$ at pixel $p_i$ on the depth map $\mathcal{D}(\tilde{\mathcal{S}}_v)$ will influence the points in $\mathcal{V}(p_i)$ via $w_{ij}$ in Eq.~\eqref{eq:depth_weight}.

\noindent{\bf Mask Loss.} 
By projecting the 3D surface to the 2D plane, we can also get its 2D binary mask $\mathcal{M}(\mathcal{P}_v)$. For each pixel on $\mathcal{M}(\mathcal{P}_v)$, the mask value is $1$ if its distance to the projection of $\mathcal{G}_{v}$ is less than a given threshold. Otherwise the value is $0$. After computing the mask of $\tilde{\mathcal{S}}$ and $\mathcal{T}$ as $\mathcal{M}(\tilde{\mathcal{S}}_{v})$ and $\mathcal{M}(\mathcal{T}_{v})$, we define the mask loss as:
\begin{equation}
\label{eq:global_mask_loss}
\mathbf{L}_{\mathrm{mask}}(\tilde{\mathcal{S}},\mathcal{T})=\mathbb{E}_{v\sim V}  \left\|\mathcal{M}(\tilde{\mathcal{S}}_v)-\mathcal{M}(\mathcal{T}_v)\right\|_1 ,
\end{equation}

The back-propagation process of global mask loss can be computed in the following way. Let $c_i$ denotes the value of pixel $p_i$ on $\mathcal{M}(\tilde{\mathcal{S}}_v)$, and $\nabla c_i$ denotes the gradient of $c_i$. 
The gradient of point $\tilde{s}_j\in\mathcal{\tilde{S}}_{v}$ can be calculated as:
\begin{align*}
\label{eq:mask_propogate}
\nabla \tilde{s}_j^z&=\sum_{p_i\in\mathcal{M}(\tilde{\mathcal{S}}_v)}\nabla c_i\cdot\frac{\exp(-\tilde{\rho}_{ij}/\tilde{\gamma})}{\sum\nolimits_{m=1}^M\exp(-\tilde{\rho}_{im}/\tilde{\gamma})},
\end{align*}
where $\nabla \tilde{s}_j^z$ denotes the gradient of z-coordinate of $\tilde{s}_j$, $\tilde{\gamma}$ controls the sharpness of this loss and $\tilde{\rho}_{ij}$ denotes the squared distance between $p_i$ and the projection (to the $x0y$ plane) of $\tilde{s_j}$. Since we adopt the soft rasterization strategy, the mask loss at one pixel can influence all points of $\tilde{\mathcal{S}}$.

\noindent{\bf As Rigid As Possible Loss.} The edge length of the deformed point cloud should be close with the original edge length via the following term:
\begin{equation}
\label{eq:smooth_loss}
\mathbf{L}_{\mathrm{arap}}(\tilde{\mathcal{S}})=\sum\limits_{(\mathbf{p}, \mathbf{q}) \in \mathcal{E}} (\|\mathbf{p} - \mathbf{q}\|_2 - d_{ij})^2,
\end{equation}
where $\mathcal{E}$ is the edge set which is constructed by KNN in the input point cloud $\mathcal{S}$, and $d_{ij}$ is the distance of vertex pair in the input point cloud.

\noindent{\bf Regularization Terms.} Rigid transformation $\psi_{k}$ includes rotation matrix $\mathbf{R}_{k}$ and translation vector $\mathbf{t}_{k}$. We constrain that the norm of translation vector as:
\begin{equation}
\label{eq:trans_loss}
\mathbf{L}_{\mathrm{tran}}(\mathbf{t}_{k})=\|\mathbf{t}_{k}\|_2^2.
\end{equation}
Considering that the non-rigid deformation may have jumps like the joints of human body, we add one sparsity term on the skinning weights by:
\begin{equation}
\label{eq:weight_loss}
\mathbf{L}_{\mathrm{sparse}}(\mathbf{w}_{k}^{k})=\| \mathbf{w}_{k}^{k} \|_1.
\end{equation}

The total loss at stage $k$ is constituted by all above terms:
\begin{equation}
\begin{split}
\label{eq:overall_loss}
\mathbf{L}^k = & \mathbf{L}_{\mathrm{depth}}(\mathcal{S}^{k},\mathcal{T}) + \beta_{1} \mathbf{L}_{\mathrm{mask}}(\mathcal{S}^{k},\mathcal{T}) \\
 & + \beta_{2} \mathbf{L}_{\mathrm{arap}}(\mathcal{S}^{k}) + \beta_{3} \mathbf{L}_{\mathrm{tran}}(\mathbf{t}_{k}) + \beta_{4} \mathbf{L}_{\mathrm{sparse}}(\mathbf{w}_{k}^{k})
\end{split}
\end{equation}

The final loss function is a combination of all stages:
\begin{equation}
\mathbf{L} = \sum_{i=1}^K \gamma^{K-i} \mathbf{L}^i,
\end{equation}
where $\gamma$ is an exponentially increasing weights for later stages.
\section{Experiments}
\label{Experiments}

In this section, we give the implementation details, ablation studies, results, and comparisons.


\subsection{Implementation Details}
\label{implementation_details}
\noindent{\bf Dataset.} We first test on a dataset including four types of deformable objects, including clothed body, naked body, cats, and dogs, with a total of 155474 training pairs and 7688 testing pairs. For the clothed and naked body data, we use the HumanMotion~\cite{DBLP:journals/tog/VlasicBMP08} and SURREAL~\cite{DBLP:conf/cvpr/Varol0MMBLS17} datasets. The cats and dogs are from the TOSCA~\cite{DBLP:series/mcs/BronsteinBK09} dataset. We also test on the FaceWareHouse dataset~\cite{DBLP:journals/tvcg/CaoWZTZ14} which contains face shapes of 150 different individuals with 47 different expressions. We randomly select $20000$ and $500$ pairs for training and testing, where each pair of source and target are different expressions of two randomly selected people. 
We also test our method on raw scanned data in DFAUST~\cite{dfaust:CVPR:2017} dataset ($2732$ training pairs and $100$ testing pairs), which contains natural noise, outliers and incompleteness. 
Moreover, to verify that our model can be generalized to the rigid registration task, we train a rigid version of our network on the ModelNet40 dataset~\cite{DBLP:conf/cvpr/WuSKYZTX15}. We split each category into $9:1$ for training and testing. Each extracted point cloud is firstly centered and then scaled into the sphere with radius $0.5$. To construct pairs, we use random rotation angles at the range of $[0,45^\circ]$ and translations at the range of $[-0.5,0.5]$. 

\noindent{\bf Implementation Details.}  {For the non-rigid registration experiments on deformable objects and human faces, the number of extracted points are $2048$ and $5334$, respectively. For the raw scanned data, we sample $2048$ control points to feed into the network and then warp the whole raw scanning model by Radial basis function interpolation.} For the rigid registration experiments, the number of points is $1024$. For FaceWareHouse data, we crop the front face from the original topology and directly take the $5334$ vertices as the point cloud. 
In the non-rigid registration experiments, the weights of each term in Eq.~\eqref{eq:overall_loss} are set as $0.1,0.01,0.1,10$ and $\gamma=1.0$. For rigid registration, we only use the first two terms with $\beta_{1}=0.1$ and $\gamma=0.8$. We adopt a warm-up training strategy to train the model. At the start, the network is trained with only $1$ recurrent stage. Every $5K$ iterations of training, we increase the number of recurrent stages by $1$ until the number of stages reaches 7. In the three experiments, the total number of training iterations are $100K$, $50K$, and $50K$, respectively, and the batch sizes are all $4$. All the training and testing is conducted on a workstation with 32 Intel(R) Xeon(R) Silver 4110 CPU @ 2.10GHz, 128GB of RAM, and four 32G V100 GPUs. For input point cloud with different numbers $1024$, $2048$ and $5334$, the inference time for 7 recurrent stages takes $0.14$s, $0.25$s and $0.55$s, respectively. The camera views of the multi-view loss terms are sampled from the spherical coordinate of the unit sphere, as shown in Fig.~\ref{fig:teaser}.

\noindent{\bf Evaluation Metrics.} We evaluate the registration performance with CD and EMD. For FaceWareHouse dataset, we further compute the point-wise mean squared error (MSE) as they share the same topologies. For rigid registration, we evaluate the transformation with root mean squared error (RMSE) and mean absolute error (MAE), as in~\cite{DBLP:conf/iccv/WangS19}.

\begin{table}[t]
	\small
	\setlength\tabcolsep{2.8pt}
	\begin{center}
		\begin{tabular}{lcc|cc}
			\bottomrule
			Loss&$\#$View &$\#$Stage   & CD & EMD   \\ \hline
			Chamfer&-&  $7$                              & 1.652        & 4.962       \\ 
			EMD&-&  $7$ & 3.241  & 4.542 \\ 
			Depth &$5^2$ &  $7$                    & 0.701        & 0.431       \\ 
			Depth+Mask&$5^2$&   $7$                   & 0.691        & 0.426       \\ 
			Depth+Mask&$7^2$&   $7$                   & 0.618        & 0.396       \\ 
			Depth+Mask&$9^2$&  $7$                    & 0.600        & 0.386       \\ 
			Depth+Mask&$11^2$&  $3$                  & 10.717        & 12.404       \\ 
			Depth+Mask&$11^2$&  $4$                   & 4.571        & 5.234       \\ 
			Depth+Mask&$11^2$&  $5$                    & 2.329        & 3.116       \\ 
			Depth+Mask&$11^2$&  $6$                    & 1.110        & 0.951       \\ 
			Depth+Mask&$11^2$&  $7$              & \textbf{0.599}        & \textbf{0.386}       \\ 
			\toprule
		\end{tabular}
	\end{center}
	\caption{Results of the ablation study, with metrics CD($\times 10^{-4}$) and EMD($\times 10^{-3}$).}
	\label{tab:ablation}
	\vspace{-5mm}
\end{table}

\subsection{Ablation Study}
We first conduct ablation studies to demonstrate the importance of each component. Specifically, we use the dataset of four categories of deformable objects as the benchmark in this part. 
The ablation studies are designed for the loss function, the number of views, and the number of recurrent stages. Tab.~\ref{tab:ablation} shows the quantitative results. As the second row shows, the registration errors of Chamfer Distance loss and Earth-Mover Distance loss are still quite large even it is already the best result we have tried with different parameters. With our proposed depth loss, the registration result is significantly improved (the $3$-rd row). The additional mask loss (the $4$-th row) brings further performance gains. We also gradually increase the number of views (the $4,5,6,11$-th rows) and finally choose $11^2$ views. With the view number fixed to $11^2$, we show the results of different recurrent stages in the $7-11$-th rows. It can be observed that the registration accuracy continues to get better with more iterations. In order to balance the registration accuracy and computation speed, we finally use 7 stages.

\begin{figure}[t]
	\centering
	\includegraphics[width=1\columnwidth]{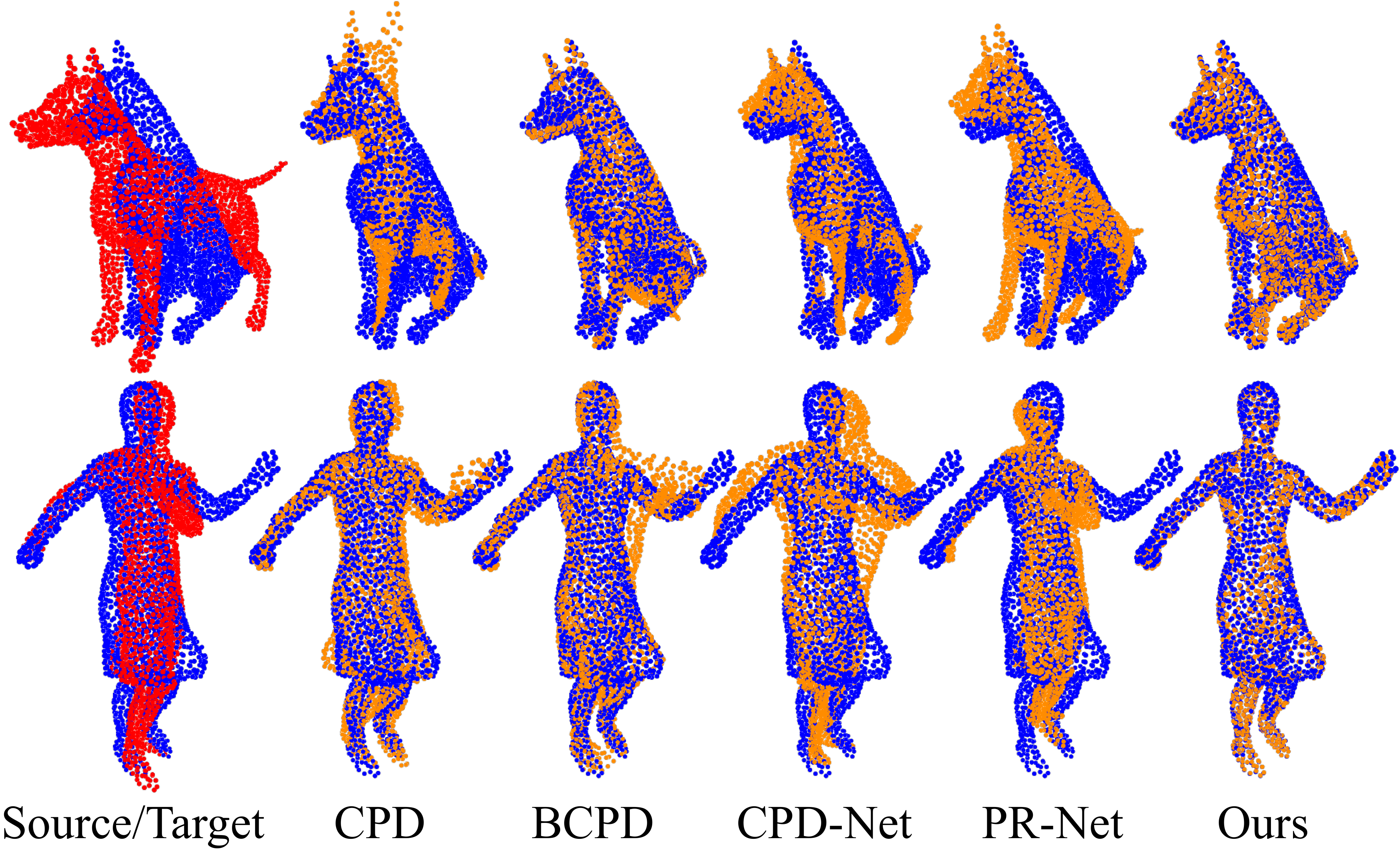}
	\caption{Comparison on the deformable objects. The source point cloud, target point cloud and deformed point cloud are visualized by red, blue and orange respectively.}
	\label{fig:comparison}
\vspace*{-2mm}
\end{figure}

\subsection{Results and Comparisons}

\subsubsection{Registration for deformable objects}
\label{non-rigid_deformable_objects}

We compare with the classic optimization method CPD~\cite{DBLP:journals/pami/MyronenkoS10}, its recently improved version BCPD~\cite{0A}, and learning based methods CPD-Net~\cite{DBLP:journals/corr/abs-1906-03039} and PR-Net~\cite{DBLP:journals/corr/abs-1904-01428}.

\begin{table}[th]
	\small
	\setlength\tabcolsep{1.8pt}
	\begin{center}
		\begin{tabular}{lccccccc}
			\toprule
			Dataset & Metric   &Input &CPD  &BCPD  &CPD-Net & PR-Net & Ours \\ 
			\midrule
			\multirow{2}{*}{Deform} &CD &37.246 &4.126 &2.375 &14.678 &29.457 &\textbf{0.599} \\ 
			&EMD &25.952 &7.853 &5.478 &21.696 &25.192 &\textbf{0.386} \\ 
			\midrule
			
			\multirow{2}{*}{Face} &EMD &1.230 &1.168 &0.979 &1.054 &1.304 &\textbf{0.578} \\ 
			&MSE &21.469 &9.568 &8.013 &13.752 &14.575 &\textbf{5.245} \\ \bottomrule
		\end{tabular}
	\end{center}
	\caption{Results on the deformable objects dataset (denoted as Deform for short, with metrics CD($\times 10^{-4}$) and EMD($\times 10^{-3}$)) and the FaceWareHouse dataset (denoted as Face for short, with metrics EMD($\times 10^{-2}$) and MSE($\times 10^{-4}$)).}
	\label{tab:data_comparison}
\end{table}

Tab.~\ref{tab:data_comparison} shows the performance of different methods on the deformable objects and the FaceWareHouse dataset. Fig.~\ref{fig:comparison} shows the qualitative comparisons. 
Our method significantly outperforms previous methods, both qualitatively and quantitatively. The optimization based methods CPD and BCPD can not handle large-scale non-rigid deformation, and thus they can not perform well on the challenging test set. CPD-Net can not handle the deformation, either, which should be caused by the large freedom of the point-wise displacement vector. Although PR-Net reduces the freedom, the spline-based representation can not express the deformation well, leading to poor results. In comparison, our method performs best thanks to our well-designed non-rigid representation, network structure, and loss terms.

\begin{figure}[t]
	\centering
	\includegraphics[width=1\columnwidth]{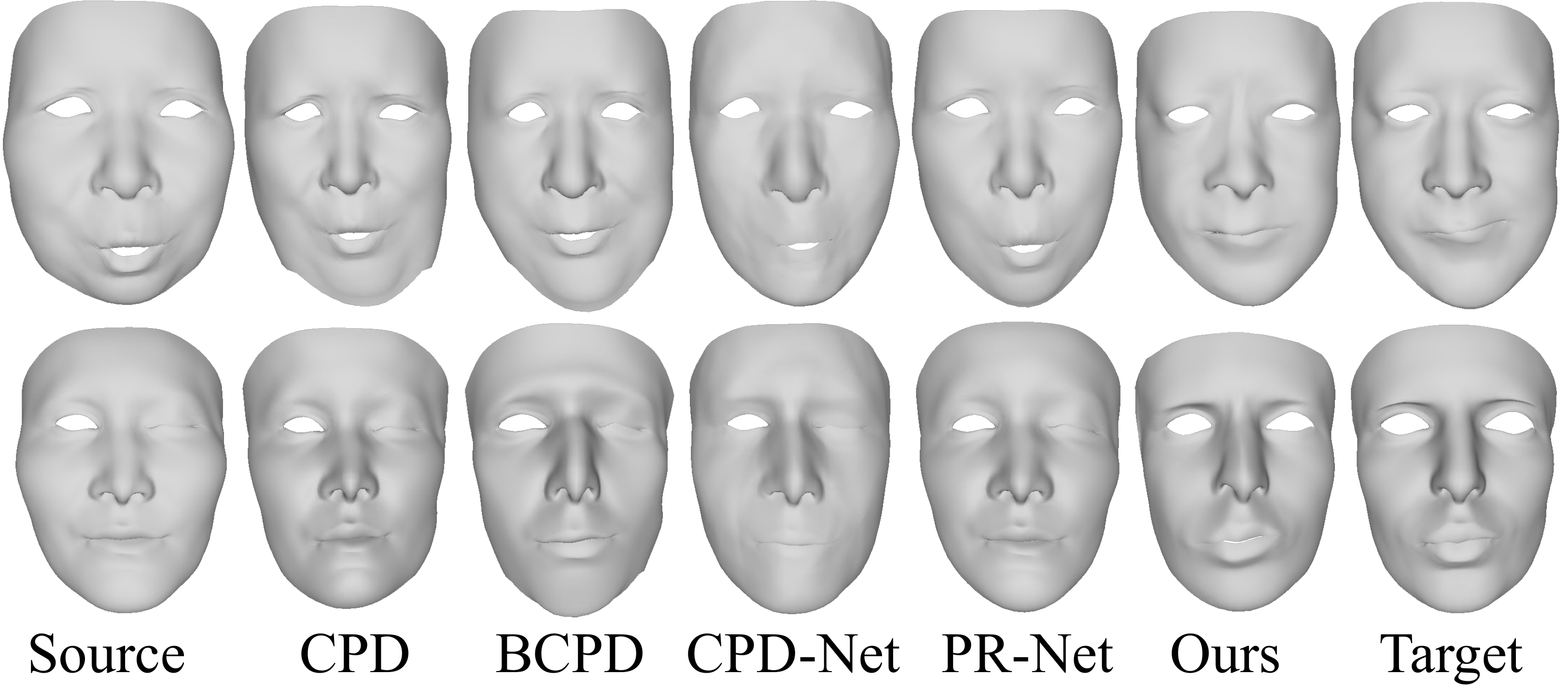}
	\caption{Comparison on FaceWareHouse dataset. For better visualization, we display the surface model with mesh connectivities.}
	\label{fig:comparison_face}
\vspace*{-5mm}
\end{figure}

\vspace{-1.5mm}
\subsubsection{Registration for human faces}
\label{face_expression}

Besides testing on coarse scale non-rigid deformation samples, we also test and compare with other methods on fine scale deformation samples. We try to register one face shape from a person with expression to another face shape from another person with a different expression. Considering that we only deal with the front face, we put all camera views on the front side of the face in this experiment.


Tab.~\ref{tab:data_comparison} shows the performance of different methods on the FaceWareHouse dataset. Fig.~\ref{fig:comparison_face} shows the qualitative comparisons. Our method performs considerably better than all previous methods, not only on EMD and MSE metrics but also on the visual quality of the registration results. In the first row of Fig.~\ref{fig:comparison_face}, our method is the only one that is able to deform the open mouth to closed. Moreover, in the second row, our result is the only one that can make the eyes open, demonstrating that our method has a better ability to capture the high-precision face deformation than previous methods.

\vspace{-1.5mm}
\subsubsection{Registration for raw scanned data}
\label{raw_scan}

\begin{figure}[t]
	\centering
	\includegraphics[width=0.9\columnwidth]{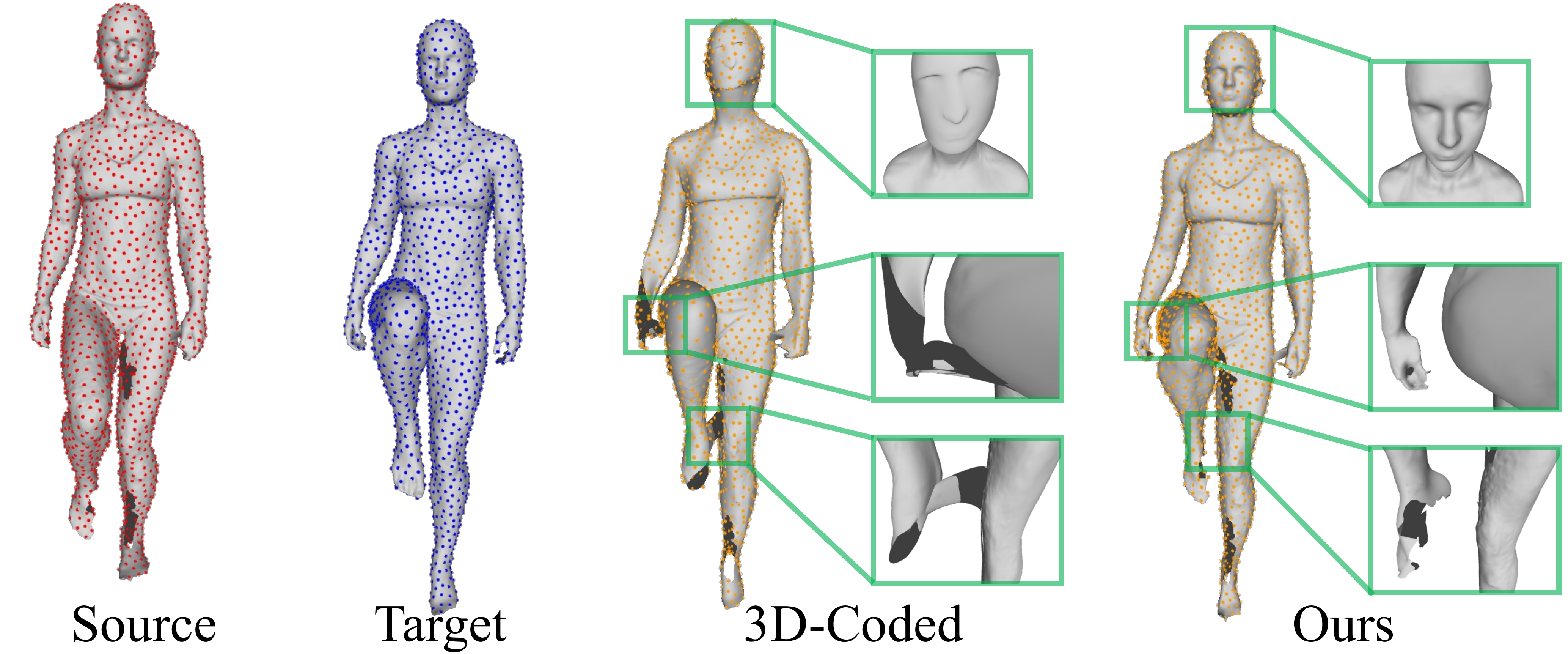}
	\caption{Comparison with 3D-Coded on the DFAUST dataset.}
	\label{dfaustfig}
\vspace*{-5mm}
\end{figure}

We also train and test on the raw scanned dataset DFaust~\cite{dfaust:CVPR:2017} and compare with 3D-Coded~\cite{DBLP:conf/eccv/GroueixFKRA18}. Fig.~\ref{dfaustfig} shows one comparison result. The average CD ($\times 10^{-4}$) distance between the testing set of the source, the 3D-Coded results and our results with the target surface are 9.02, 0.43, 0.24, respectively. Although some defects are contained in this dataset, our method still achieves satisfactory results and better preserves the geometric structures and details than 3D-coded~\cite{DBLP:conf/eccv/GroueixFKRA18}. This experiment shows that our method still works well for the real scanned data.

\vspace{-1.5mm}
\subsubsection{Rigid registration}
\label{rigid_registration}

Our full framework can also be easily extended to perform the rigid registration task. In this experiment, we convert our network into a rigid version by predicting a single rigid transformation at each stage (as discussed in Sec.~\ref{sec:representation}) and test the performance on the ModelNet40 dataset. 


The comparison methods include local optimization based method ICP~\cite{DBLP:journals/pami/BeslM92}, 
two global optimization based method Go-ICP~\cite{DBLP:journals/pami/YangLCJ16} and FGR~\cite{DBLP:conf/eccv/ZhouPK16}, and learning based method DCP~\cite{DBLP:conf/iccv/WangS19}. DCP~\cite{DBLP:conf/iccv/WangS19} is supervised by the ground truth rotation and translation, and trained with the same dataset in our own network. For a fair comparison with DCP, we also train another variant of our RMA-Net that uses ground truth as supervision except training a model in an unsupervised manner (same as previous experiments). Our model is trained with 7 stages, but can be used with arbitrary stages at inference time. In our experiments, we use 10 stages during testing. Tab.~\ref{tab:same_category} shows the rigid registration performance of different methods on ModelNet40 dataset. It can be observed that our method performs better than previous methods in the rigid case, which demonstrates the general applicability of our proposed framework.




\begin{table}[th]
	\small
	\setlength\tabcolsep{1.8pt}
	\begin{center}
		\begin{tabular}{lcccc}
			\toprule
			Metric   & RMSE(\textbf{R}) & MAE(\textbf{R}) & RMSE(\textbf{t}) & MAE(\textbf{t}) \\ \midrule
			ICP~\cite{DBLP:journals/pami/BeslM92}         & 19.041        & 7.585       & 0.133        & 0.154       \\ 
			Go-ICP~\cite{DBLP:journals/pami/YangLCJ16}  & 13.086        & 1.891       & 0.060        & 0.026       \\ 
			FGR~\cite{DBLP:conf/eccv/ZhouPK16}  & 10.143        & 1.928       & 0.048        & 0.030       \\ 
			DCP~\cite{DBLP:conf/iccv/WangS19}             & 2.057        & 1.313       & 0.013        & 0.023       \\ \midrule
			Ours (unsupervised)                                          & 1.287        & 0.344       & 0.008        & \textbf{0.007}       \\ 
			Ours (supervised)                                         & \textbf{0.735}        & \textbf{0.265}       & \textbf{0.006}        & 0.009       \\ 
			\bottomrule
		\end{tabular}
	\end{center}
	\caption{Comparison on the ModelNet40 dataset.}
	\label{tab:same_category}
\vspace*{-5mm}
\end{table}


\section{Conclusion}
\label{Conclusion}

We have presented RMA-Net, an unsupervised learning framework for non-rigid registration. The main contributions of RMA-Net lie in two aspects. First, We propose a new non-rigid representation, which is learned with a recurrent network. Second, we designed a multi-view alignment loss function to guide the network training without ground truth correspondence as supervision. Extensive ablation studies have verified the effectiveness of each component in our full framework. We also outperform previous state-of-the-art non-rigid registration methods by a large margin, demonstrating the superiority of our proposed method.

\small {\noindent{\bf Acknowledgement} This work was supported by the Youth Innovation Promotion Association CAS (No. 2018495), and ``the Fundamental Research Funds for the Central Universities''.}

{\small
\bibliographystyle{ieee_fullname}
\bibliography{egbib}
}

\section*{Appendix}
This supplementary material provides more details that were not given in the main paper due to space constraints, including the correlation computation in the network, the construction of the deformable objects dataset, more experimental results, comparisons and analysis.

\subsection*{Correlation Computation}
\begin{figure}[ht]
	\centering
	\includegraphics[width=1\columnwidth]{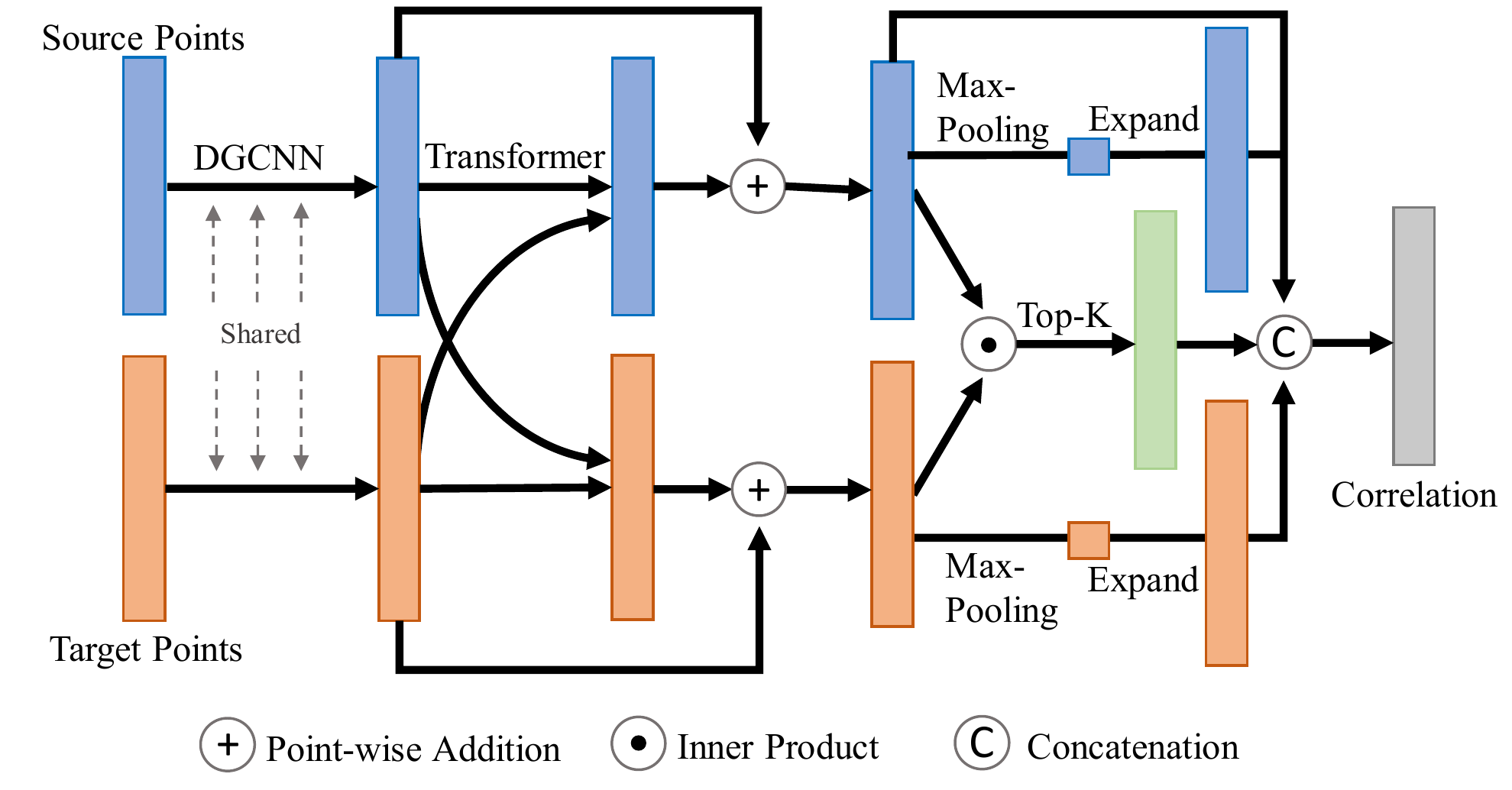}
	\caption{The process of extracting the correlation from the deformed point cloud and the target point cloud.}
	\label{fig:correaltion}
\end{figure}

In each recurrent stage, we extract features from the current deformed point cloud and the target point cloud, and calculate the correlation (\ie, dot-product) to measure the gap between them. This process is illustrated in Fig.~\ref{fig:correaltion}. The features are extracted with DGCNN~\cite{DBLP:journals/tog/WangSLSBS19} and Transformer~\cite{DBLP:conf/nips/VaswaniSPUJGKP17}, where the DGCNN has $5$ $EdgeConv$ layers, and the Transformer has $4$ heads in multi-head attention. The feature channel number is set as $1024$. A top-K operation is then performed on the last dimension of the computed correlation to make the output independent on the number of points, where the K is set as $1024$. Then we concatenate the top-K result and the deep features of two point clouds together as the input to next iterative update.

\subsection*{Deformable Objects Dataset}
For the clothed body data, we use the HumanMotion~\cite{DBLP:journals/tog/VlasicBMP08} dataset, which contains $10$ Human motion sequences. We randomly extract $37600$ training pairs and $1652$ testing pairs. For the naked body data, we use the SURREAL~\cite{DBLP:conf/cvpr/Varol0MMBLS17} dataset, which is composed of $68036$ videos containing synthetic human bodies represented by SMPL~\cite{DBLP:journals/tog/LoperM0PB15} model. We extract $65000$ and $3036$ pairs for training and testing, respectively. The cats and dogs are from the TOSCA~\cite{DBLP:series/mcs/BronsteinBK09} dataset. Because the amount of data is not large ($11$ cats and $9$ dogs), we construct a large amount of data through ACAP~\cite{DBLP:journals/corr/abs-1709-01250} interpolation. For training, we construct $20604$ dog pairs and $32270$ cat pairs. For testing, we construct $1500$ dog pairs and $1500$ cat pairs. Overall, we use $155474$ training pairs and $7688$ testing pairs in this experiment.

\subsection*{More Results and Comparisons}
\begin{figure*}[ht]
	\centering
	\includegraphics[width=2\columnwidth]{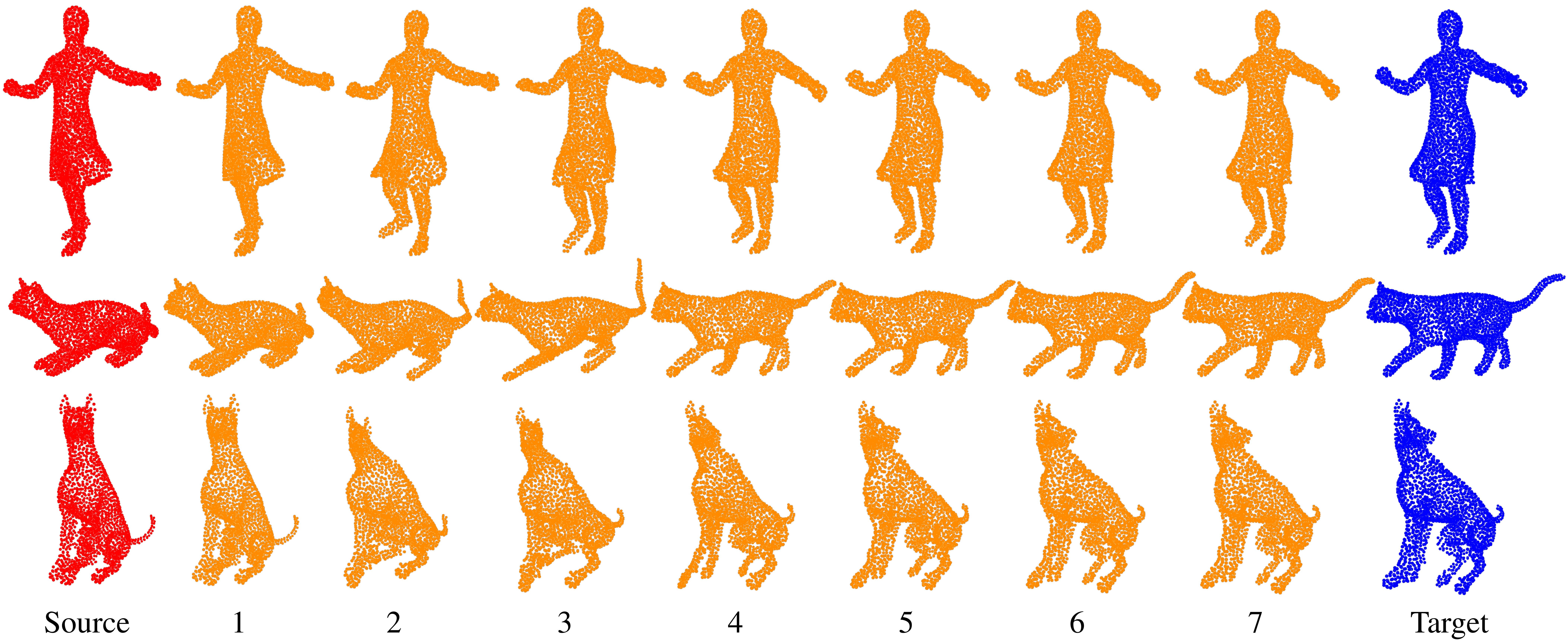}
	\caption{The results of different stages.}
	\label{fig:stages}
\end{figure*}

\noindent{\bf Results of Different Stages.} 
To show the effect of our recurrent strategy more clearly, we visualize some samples from the deformable objects dataset at each stage. As illustrated in Fig.~\ref{fig:stages}, the deformed point clouds become closer and closer to the target point cloud during the recurrent process.

\begin{figure*}[ht]
	\centering
	\includegraphics[width=2.0\columnwidth]{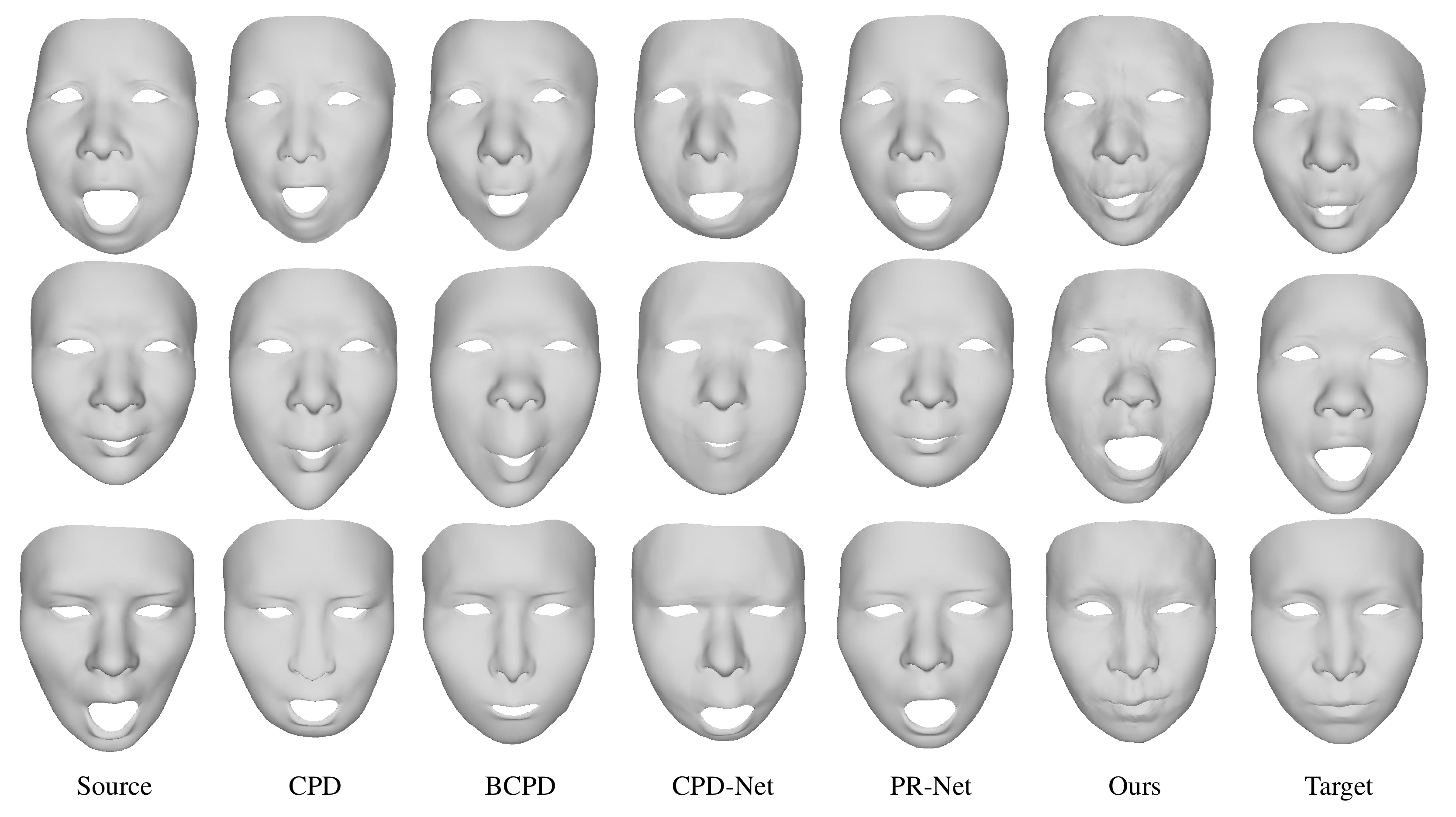}
	\caption{Comparison on testing samples in the FaceWareHouse dataset.}
	\label{fig:face_supp}
\end{figure*}

\begin{figure*}[ht]
	\centering
	\includegraphics[width=1.7\columnwidth]{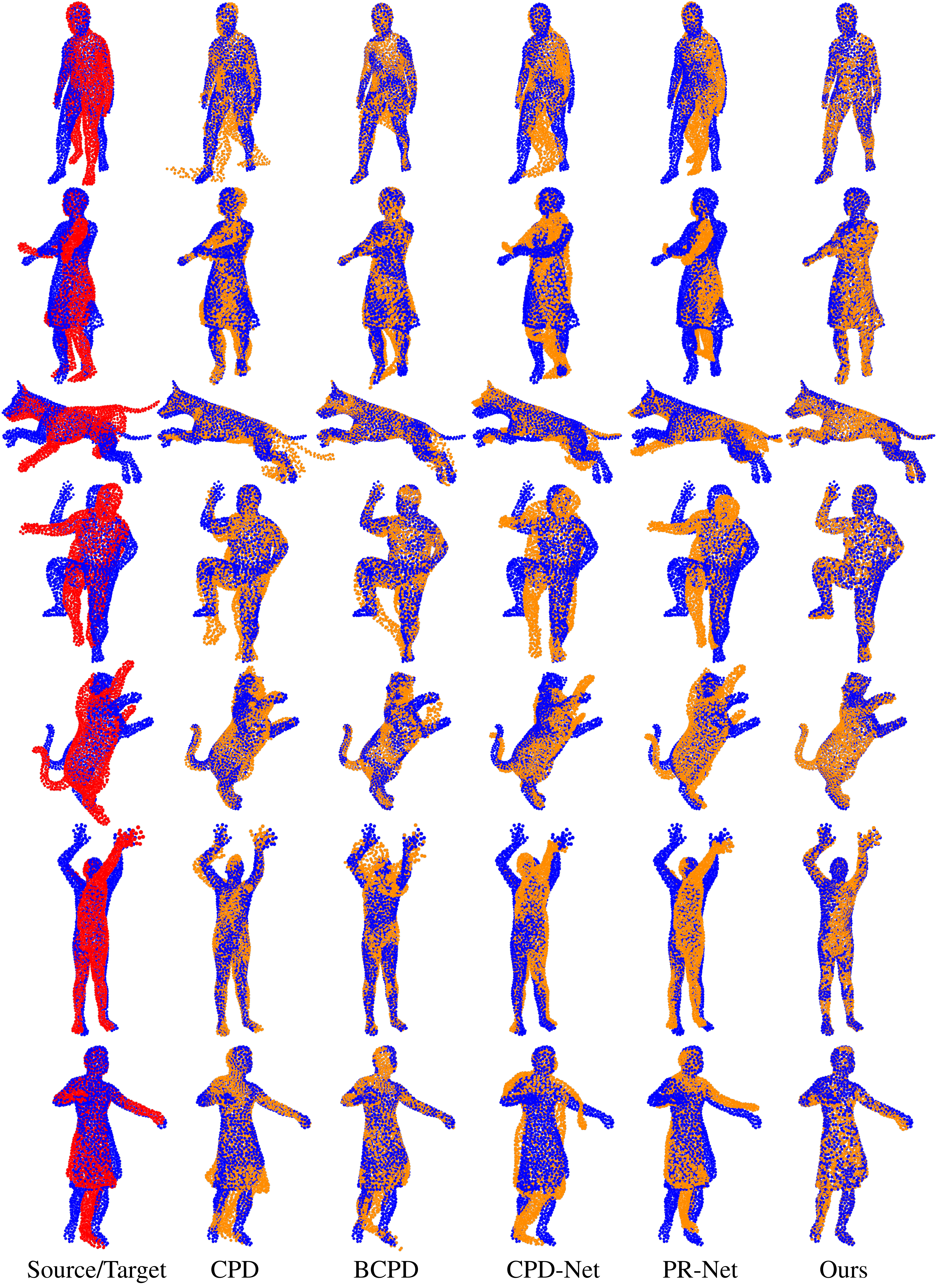}
	\caption{Comparisons on testing samples in the deformable objects dataset.}
	\label{fig:comparisons}
\end{figure*}

\noindent{\bf Comparison on More Samples.} 
We also show more samples to compare our method with CPD~\cite{DBLP:journals/pami/MyronenkoS10}, BCPD~\cite{0A}, CPD-Net~\cite{DBLP:journals/corr/abs-1906-03039}, and PR-Net~\cite{DBLP:journals/corr/abs-1904-01428}. In Fig.~\ref{fig:comparisons}, we show more comparison samples in the deformable objects dataset. In Fig.~\ref{fig:face_supp}, we show more comparison samples in the FaceWareHouse~\cite{DBLP:journals/tvcg/CaoWZTZ14} dataset.  

\begin{figure*}[ht]
	\centering
	\includegraphics[width=2\columnwidth]{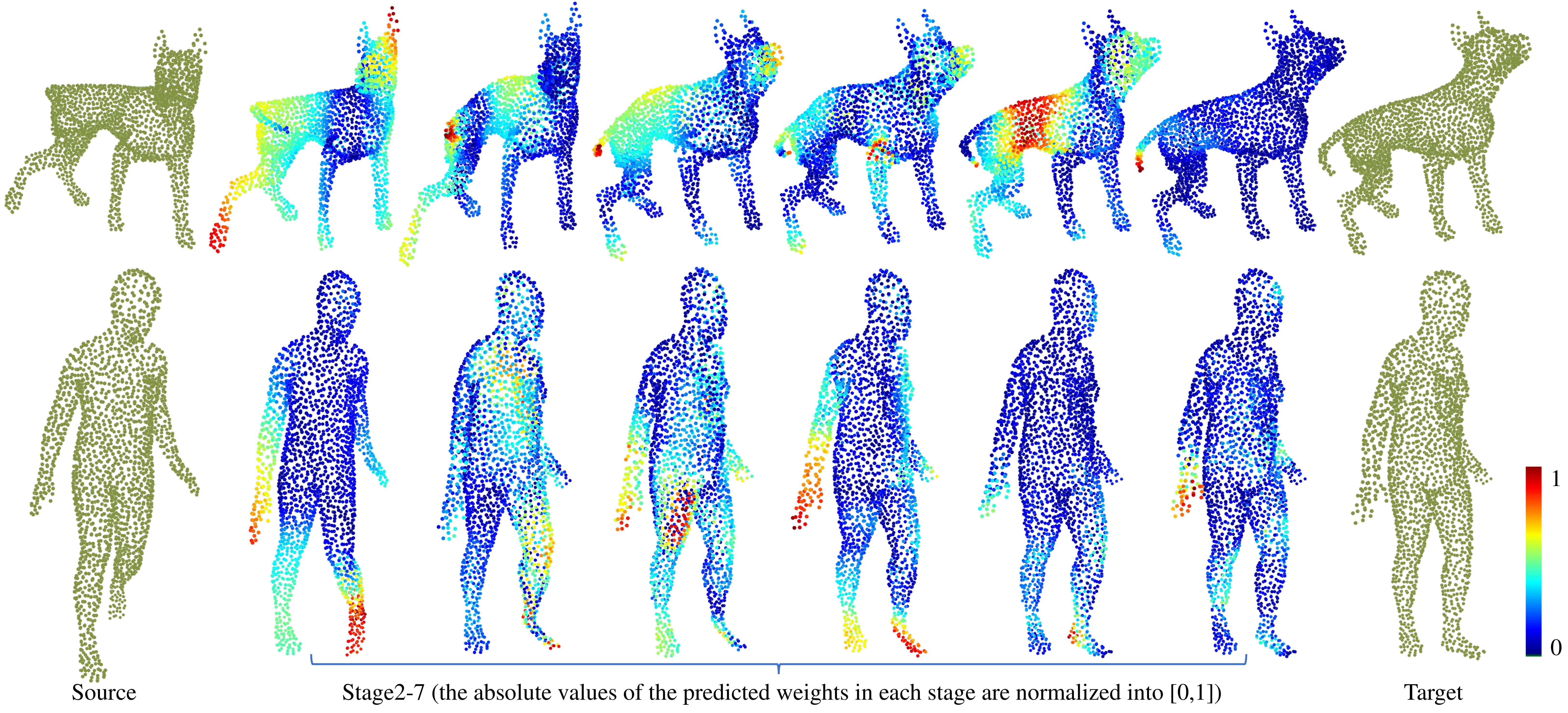}
	\caption{The weight distribution on the deformed surface.}
	\label{fig:analysis_w}
\end{figure*}

\subsection*{Distribution of the Skinning Weights}
To observe the weight distribution on the deformed surface better, we select some samples and visualize the skinning weights in Fig.~\ref{fig:analysis_w}. We first show the deformation result point clouds of each stage. Among them, we visualize the weight of stage $2$ to $7$ ($\{\mathbf{w}_{k}^{k}\}$, k=$2, \cdots ,7$). We normalize the absolute of each $\mathbf{w}_{k}^{k}$ into range $[0,1]$ and render the points according to the color bar. From Fig.~\ref{fig:analysis_w}, we can see that the distribution of the skinning weights is smooth and local. The smoothness of the skinning weight distribution helps the resulting point clouds maintain a reasonable shape. The locality of the weight distribution verifies that our recurrent strategy reduces the freedom of each stage. Moreover, we can observe that the weights tend to concentrate on the parts with relatively large deformation, which also explains the effectiveness of our method for large-scale deformations.

\end{document}